\documentclass[11pt]{article}

\usepackage[preprint]{acl}
\usepackage{booktabs}
\usepackage{threeparttable}
\usepackage{fontspec}
\usepackage{amsmath}
\usepackage{multirow}
\usepackage{times}
\usepackage{latexsym}
\usepackage{graphicx}
\usepackage[notransparent]{svg}
\usepackage{tikz}
\usetikzlibrary{trees}

\usepackage[T1]{fontenc}

\usepackage{float}

\usepackage{graphicx}

\usepackage[english]{babel}

\babelprovide[import]{hindi}

\usepackage[english]{babel}

\newfontfamily\hindifont{NotoSansDevanagari-Regular.ttf}[
    Script=Devanagari, 
    Scale=MatchLowercase, 
    Path=./,              % Looks in the current directory
]

\newfontfamily\ipafont[Path=./,Extension=.ttf]{CharisSIL-Regular}

\usepackage{xcolor}
\usepackage{enumitem}
\usepackage{tabularx}

\usepackage[verbose=silent]{microtype}

\usepackage{inconsolata}

\usepackage{hyperref}
\usepackage{amssymb}

\title{DonorRank: Donor Language Selection \\ for Low-Resource Cross-Lingual Speech Recognition}
\author{Akriti Dhasmana \and Aarohi Srivastava \and David Chiang \\
    Computer Science and Engineering \\
 University of Notre Dame \\
 Notre Dame, IN, USA \\ \texttt{\{adhasman, asrivas2, dchiang\} @nd.edu}}

\usepackage{comment}

\begin{document}

\maketitle

\begin{abstract}
    Low-resource automatic speech recognition (ASR) often relies on cross-lingual transfer from higher-resource languages due to limited availability of transcribed speech. However, selecting an effective donor language is a nontrivial issue, particularly in settings with spontaneous speech, linguistic variation, and uneven resource availability. We present DonorRank, a learning-to-rank framework that uses linguistic and dataset-level features along with previously observed transfer outcomes to predict promising donor languages for zero-shot ASR. We evaluate DonorRank on two multilingual speech corpora of Indic and African languages. DonorRank produces informative donor rankings across both collections, outperforming common heuristics in the more closely related Indic setting while remaining competitive in the more linguistically diverse African setting. Our analysis reveals that effective donors are not always the closest linguistic relatives and that the most informative predictors vary across language collections. These findings highlight the value of learned donor selection for prioritizing transfer experiments in low-resource ASR. We will also release all of our evaluation results, feature matrices for all languages, scripts for training the DonorRank framework, and pairwise transfer performance matrices for our datasets.
\end{abstract}

\section{Introduction}

Building automatic speech recognition (ASR) systems for the world's thousands of languages is challenging since most languages lack sufficient transcribed speech for supervised training. Cross-lingual transfer has emerged as a common strategy for low-resource ASR, where a model is fine-tuned on one or more higher-resource \emph{donor} languages before being applied to an unseen \emph{target} language. However, performance depends critically on the choice of donor languages, and may vary substantially even among closely related languages. Identifying an effective donor is nontrivial and often requires evaluating multiple candidates on the target language, a computationally expensive, sometimes impractical process for low-resource settings. This raises the question: how can we efficiently identify promising donor languages for a new target without evaluating every candidate?
 
Cross-lingual transfer in low-resource ASR has additional challenges beyond those in multilingual NLP. Unlike the benchmarks built on standardized text or read speech, low-resource ASR often relies on spontaneous speech collected in real-world settings, with variable pronunciation, speaking style and recording conditions. This is compounded by the fact that many low-resource languages have limited orthographic standardization making lexical similarity less reliable. Consequently, effective donor selection may require considering multiple linguistic and dataset-specific factors beyond genetic relatedness or lexical similarity alone. 

To address this, we introduce \emph{DonorRank}, a learning-to-rank framework that predicts promising donor languages for zero-shot ASR using linguistic and dataset-level features. By learning from previously observed transfer relationships, DonorRank can rank candidate donors for unseen target languages without requiring exhaustive pairwise transfer experiments. Rather than relying solely on genealogical relatedness or resource availability, DonorRank combines multiple sources of information to identify promising transfer partners.

We evaluate DonorRank on two multilingual speech collections, a subset of the VAANI corpus \cite{vaani2025} with Devanagari-script Indic language varieties (which we call VAANI-D),  and the WAXAL corpus \cite{diack2026waxallargescalemultilingualafrican}, with African languages from multiple language families and writing systems.

Our study makes the following contributions:
\begin{enumerate}
    \item We introduce DonorRank, a learning-to-rank framework for selecting promising donor languages for zero-shot ASR across diverse low-resource multilingual settings.
    \item We show that DonorRank learns informative rankings even with limited transfer observations and generalizes to unseen donor languages, substantially outperforming baselines in the Indic language setting.
    \item We identify linguistic and dataset-level factors associated with successful transfer, showing that effective donors are not always the closest genetic relatives.
\end{enumerate}

\section{Related Work}

\paragraph{Donor Language Selection in Text}
Selecting donor languages for transfer learning in low-resource settings has traditionally relied on heuristics based on genetic relatedness or linguistic intuition. To move beyond such methods, \citet{lin19acl} introduced LangRank, a learning-based framework for \textit{text processing} that predicts transfer performance using typological, inventory, and geographic features from the URIEL/lang2vec database~\cite{littell-etal-2017-uriel} along with dataset specific features. Similarly, NN-Rank~\cite{ebrahimi-etal-2025-model} uses learned dataset embeddings for ranking source languages in text-based NLP tasks. Subsequent text processing work has examined how linguistic features influence transfer \cite{rice-etal-2025-untangling}. However, extending these approaches to low-resource ASR introduces additional challenges due to acoustic variability and dataset-specific factors.

\paragraph{Donor Language Selection in Speech}
In speech recognition, prior work has explored donor selection using data-driven similarity measures. These include methods based on acoustic token distributions ~\cite{san-etal-2024-predicting, mitsumori2025crosslingualdataselectionusing}, as well as approaches that leverage spoken language identification models to filter training data~\cite{10848811}. \citet{san-etal-2024-predicting} have explored the use of dataset-specific and linguistic features for predicting ideal candidate languages for continued pre-training (CPT) to improve downstream ASR performance on low-resource Indic languages; however, this typically requires processing large amounts of audio data and offers limited interpretability. While prior work has explored data-driven donor selection for ASR, these approaches often address different transfer settings, such as continued pre-training (CPT) or training-data selection. Unlike \citet{san-etal-2024-predicting}, who perform CPT on donor-language data followed by fine-tuning on one hour of target-language speech, we fine-tune exclusively on donor-language data and evaluate zero-shot ASR without any target-language adaptation.

\paragraph{Linguistic Predictors of Transfer}  Recent work  \cite{de-vries-etal-2022-make, cross-lingual-transfer} has explored the impact of linguistic features such as lexical-phonetic distances, language family belonging, and shared writing systems on the success of cross-lingual transfer for several text-based NLP tasks. However, the role of these features in cross-lingual ASR transfer remains less extensively studied. \citet{florian2026evaluatingeffectlinguisticrelatedness} pursue this direction in ASR, studying the impact of feature similarity in cross-lingual transfer with a two-stage fine-tuning adaption using both the donor and target language data for ASR within ten African languages. Our analysis spans a larger set of languages (19 African languages and 20 Indic languages) in a zero-resource setting by fine-tuning solely on the donor language to study the impact of linguistic and dataset feature similarity in cross-lingual zero-resource ASR transfer.

\paragraph{Our Work} In contrast, we focus on zero-shot ASR transfer to low-resource language varieties and dialects, where data is often sparse, noisy, and drawn from spontaneous speech. Our study focuses on Indo-Aryan, Niger-Congo, Afro-Asiatic, and Nilo-Saharan languages that are linguistically diverse and underrepresented in ASR and NLP resources. Despite having large numbers of speakers, many of these varieties are absent from standard ASR benchmarks, making them a compelling testbed for evaluating transfer methods in realistic low-resource settings. Inspired by LangRank~\cite{lin19acl}, our approach, \textbf{DonorRank}, uses linguistic and dataset-level features in a lightweight ranking framework, enabling both efficient donor selection and interpretable analysis of factors influencing transfer. To assess the benefits of learned donor selection in this zero-shot setting, we compare DonorRank against practical heuristics based on genealogical relatedness and resource availability.
% Because existing speech-based donor-selection methods typically target different adaptation settings or require additional acoustic representations, we evaluate DonorRank against practical donor-selection heuristics based on genealogical relatedness and resource availability.

\section{Data and Languages}

We evaluate donor selection on two complementary speech corpora: \textbf{VAANI-D}, a Devanagari-script subset of the VAANI corpus~\cite{vaani2025} comprising closely related Indo-Aryan varieties, and \textbf{WAXAL}~\cite{diack2026waxallargescalemultilingualafrican}, spanning typologically diverse African languages across multiple families and scripts spoken in Sub-Saharan Africa.

\paragraph{VAANI-D}
VAANI is a large-scale corpus ($\sim$150k hours, 10\% transcribed) of spontaneous Indic speech collected from more than 156,000 speakers across 773 districts of India. We focus on \textbf{VAANI-D}, a Devanagari-script subset of 20 Indo-Aryan varieties. We evaluate transfer across 20 donor and 18 target languages, excluding Sadri (sdr) and Surgujia (sgj) from target evaluation due to limited test data. VAANI-D provides a linguistically controlled setting for studying transfer within a dense network of related languages. It features spontaneous, code-mixed speech with variation in orthography and pronunciation across closely related varieties, making donor selection based solely on script or lexical similarity challenging. 
% It features spontaneous, code-mixed speech with limited orthographic standardization (e.g., dialects of Hindi such as Bundeli, Awadhi and Khariboli that use micro-variations in orthography and pronunciation) ensuring that donor selection can't rely solely on script or lexical similarity.

\paragraph{WAXAL}
We contrast VAANI-D with the ASR component of \textbf{WAXAL}, with $\sim$1,250 hours of transcribed spontaneous speech across 19 African languages. WAXAL has multiple language families (e.g., Bantu, Semitic) and scripts (e.g., Latin, Ge'ez), including agglutinative (e.g., Luganda, Lingala) and tonal languages (e.g., Ewe, Ikposo, Acholi). The collection also includes languages with varying degrees of orthographic standardization, reflecting another challenge for low-resource ASR. This is a strong exploratory dataset for ranking across various language families where donor language selection must generalize beyond closely related varieties.

 VAANI-D and WAXAL allow us to evaluate whether DonorRank can identify promising donors across different multilingual settings, while examining how the relative usefulness of linguistic and dataset-level features varies with the composition of the candidate language set.

\section{DonorRank}
\begin{figure}%[tbp]
    \centering
    \includegraphics[height=3.75cm, width=\columnwidth]{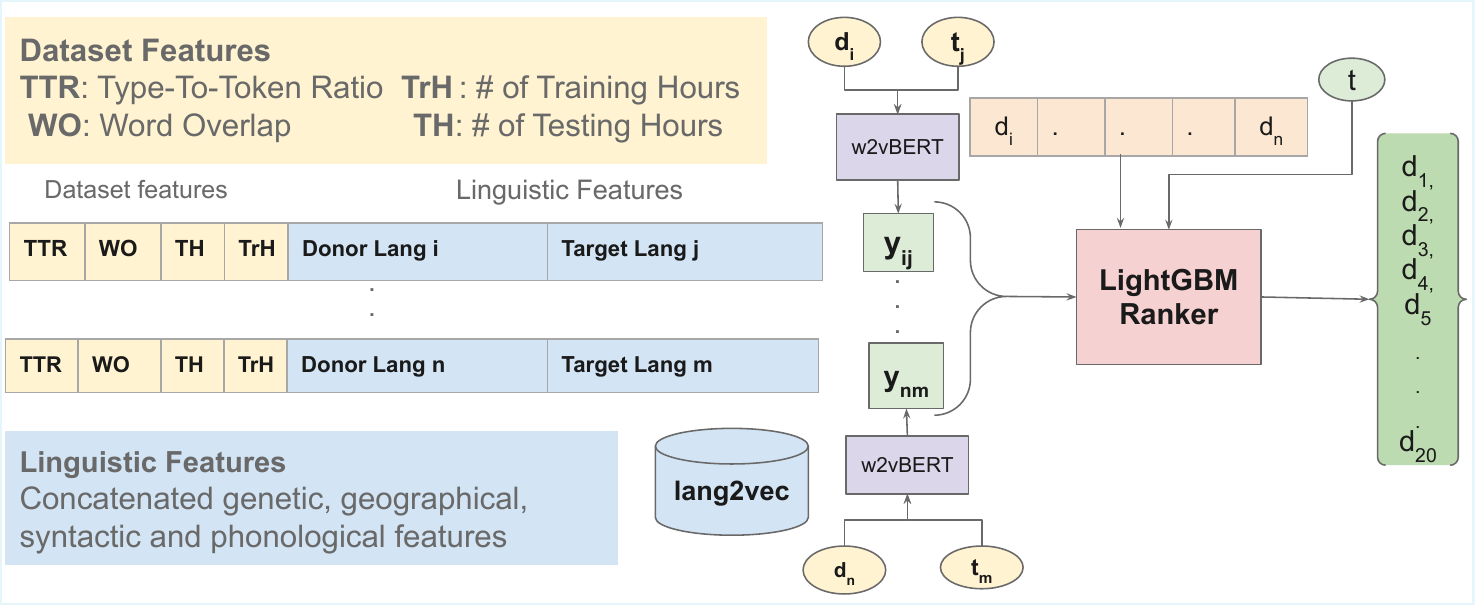}
    \caption{Overview of DonorRank. We assemble feature-based language vectors using a composition of features for donor languages $d_i \in D$ and target language $t$ to input to the ranker, which is supervised by transcription error rates from \texttt{w2vBERT}. Once trained, DonorRank predicts donor rankings for new target languages.
    }
    \label{fig:methodology}
    \vspace{-4mm}
\end{figure}
Figure~\ref{fig:methodology} illustrates the DonorRank framework which extends the LangRank framework~\cite{lin19acl}. 
\subsection{Learning Donor Rankings}
To train DonorRank, we first construct ground-truth donor rankings from pairwise transfer experiments on previously observed languages. For each donor–target pair, we fine-tune a multilingual ASR model on donor-language speech and evaluate it zero-shot on the target language. To facilitate fair comparison across donor languages, we cap the amount of fine-tuning data at 7 hours per language and use 1 hour of test data per language. Some languages have fewer available hours. Amount of training data used for each language is reported in Table~\ref{tab:lgbm-rankings-cer-only}.
We choose w2vBERT \cite{chung2021w2vbertcombiningcontrastivelearning} based on preliminary experiments and related work \cite{analysis-asr-models}. The resulting transcription error rates are an empirical measure of transfer quality, allowing donor languages to be ranked independently. These experiments are the training data for DonorRank, which learns to predict donor rankings from linguistic and dataset-level features for new target languages.

Using these rankings as supervision, we follow \citet{lin19acl} in training a LightGBM \cite{ke2017lightgbm} learning-to-rank model with the LambdaRank objective. Given a target language and a set of candidate donors, the model learns to rank donors according to their observed transfer performance using linguistic and dataset-level features. DonorRank focuses on predicting effective donor languages rather than modifying the underlying ASR model.

\subsection{Linguistic and Dataset Features}

Each donor-target pair is represented with a combination of linguistic and dataset-level features, shown in Figure~\ref{fig:methodology}. This enables the ranking model to capture various features that may influence cross-lingual transfer. The linguistic features are obtained from URIEL/lang2vec~\cite{littell-etal-2017-uriel} and include genetic, geographic, syntactic, and phonological information derived from WALS \cite{wals}, SSWL \cite{Collins2009SyntacticSO}, and PHOIBLE \cite{phoible}. Additionally, we incorporate dataset-specific features: donor training hours, target training and test hours, target type-to-token ratio, and lexical overlap between donor and target transcriptions. Depending on the languages considered as donors, different features are expected to be relevant, which we explore in our analyses. DonorRank can also be trained using subsets of the available features, allowing donor selection under different levels of linguistic and dataset information.

\subsection{Donor Ranking and Prediction}

At inference time, DonorRank uses linguistic and dataset-level features to rank candidate donors for a target language. We then evaluate the selected donors by fine-tuning an ASR model on their speech and measuring zero-shot performance on the target. We evaluate DonorRank using leave-one-target-language-out cross-validation.  For each fold, we exclude all training pairs containing the held-out language, whether as donor or target, and predict donor rankings for that language as an unseen target. This evaluation protocol assesses if DonorRank can generalize its learned donor-selection patterns to previously unseen target languages. Because the ranking model is never trained on transfer results involving the evaluation language, successful ranking shows generalization beyond previously observed language pairs.

\section{Experimental Setup}
\label{sec:experiments}
Our experiments address the following questions: 
\begin{enumerate}[itemsep=0pt, parsep=0pt]
    \item Can DonorRank accurately predict effective donor languages?
    \item Does learned donor selection improve zero-shot ASR performance?
    \item What linguistic and resource-related factors relate to successful donor selection?
\end{enumerate}

\paragraph{General Setup}
All experiments use the same pre-trained w2vBERT-2.0 checkpoint\footnote{\url{https://huggingface.co/facebook/w2v-bert-2.0} \cite{communication2023seamlessmultilingualexpressivestreaming}} and fine-tuning hyperparameters across donor languages and datasets. Models are fine-tuned for 10 epochs with a batch size of 16 and a learning rate of $5\times10^{-5}$.

\paragraph{Ranking Evaluation}
We evaluate donor ranking quality using Normalized Discounted Cumulative Gain (NDCG), which measures how effectively a predicted ranking prioritizes donors with high empirical transfer performance \cite{10.1145/582415.582418}. We define donor relevance scores as the difference between the maximum observed error rate and the donor’s WER or CER, so that lower error rates receive higher relevance scores, which are in turn used to calculate Discounted Cumulative Gain (DCG). $\text{NDCG}_k = \frac{\text{DCG}_k}{\text{IDCG}_k}$. Here, ${\text{IDCG}_k}$ is the DCG for the ideal ranking.
In addition to NDCG over the full ranking, we report NDCG@1 and NDCG@3 to assess the quality of the top-ranked donor and the top three recommendations, respectively.

\paragraph{Zero-Shot Transfer Evaluation}
To evaluate the scores obtained by the selected donor, we fine-tune w2vBERT on the highest-ranked donor language and evaluate zero-shot ASR on the target language. Performance is measured using character error rate (CER) and word error rate (WER) using Levenshtein distance between predicted and reference transcriptions. We compare DonorRank against two practical donor-selection baselines: the most genetically similar donor and a fixed high-resource donor for each dataset.

\section{Results}
We evaluate DonorRank's ability to predict donor languages, and examine if improved rankings translate into downstream ASR gains. We analyze the linguistic factors for successful transfer before concluding with a study of multi-donor transfer. In Table \ref{tab:ndcg_scores}, we compare DonorRank framework against a simple random baseline and a genetic heuristic. For the genetic heuristic, we rank donor languages using their cosine similarity to the target language based on genetic vectors obtained from the lang2vec \cite{littell-etal-2017-uriel} dataset. DonorRank outperforms both baselines on VAANI-D under WER- and CER-based ranking. On WAXAL, it slightly outperforms genetic similarity under WER-based ranking but performs worse under CER-based ranking (Table 1). These differences motivate our analysis of donor-selection signal across the two collections. 

\begin{table}[htbp]
    \centering \footnotesize
    \setlength{\tabcolsep}{2pt} % Reduces column padding to fit 7 columns cleanly
    \begin{tabularx}{\columnwidth}{l *{6}{>{\centering\arraybackslash}X}}
        \toprule
        & \multicolumn{2}{c}{\textbf{\textit{DonorRank}}} & \multicolumn{2}{c}{\textbf{Random}} & \multicolumn{2}{c}{\textbf{Genetic}} \\
        \cmidrule(lr){2-3} \cmidrule(lr){4-5} \cmidrule(lr){6-7}
        \textbf{Dataset} & \textbf{WER} & \textbf{CER} & \textbf{WER} & \textbf{CER} & \textbf{WER} & \textbf{CER} \\ 
        \midrule
        \multicolumn{7}{c}{\textbf{\textit{Mean NDCG@1}}} \\
        \midrule
        WAXAL & 82.3 & 79.5 & 32.7 & 49.7 & \textbf{95.2} & \textbf{98.1} \\
        VAANI & \textbf{76.7} & \textbf{93.0} & 47.7 & 67.8 & 60.7 & 73.2 \\
        \midrule
        \multicolumn{7}{c}{\textbf{\textit{Mean NDCG@3}}} \\
        \midrule
        WAXAL & 85.7 & 82.9 & 44.8 & 58.4 & \textbf{92.2} & \textbf{93.6} \\
        VAANI & \textbf{80.2} & \textbf{94.9} & 51.6 & 69.6 & 61.0 & 71.5 \\
        \midrule
        \multicolumn{7}{c}{\textbf{\textit{Overall Mean NDCG}}} \\
        \midrule
        WAXAL & \textbf{94.8} & 93.2 & 74.8 & 83.1 & 94.1 & \textbf{96.7} \\
        VAANI & \textbf{92.6} & \textbf{98.4} & 80.2 & 88.6 & 84.4 & 90.0 \\
        \bottomrule
    \end{tabularx}
    \caption{Mean NDCG evaluation results of predicted donor rankings across DonorRank, Random Baseline, and Genetic Similarity on WAXAL and VAANI. WER and CER indicate the transfer-performance metric used to predict ranking. Best performance per dataset and metric is in \textbf{bold}.}
    \label{tab:ndcg_scores}
\end{table}

\subsection{DonorRank produces informative rankings across datasets.}

Across both VAANI-D and WAXAL, DonorRank achieves high NDCG scores, indicating that linguistic and dataset-level features can be used to prioritize some donor languages (see Table~\ref{tab:ndcg_scores}). \textbf{On VAANI-D, DonorRank substantially outperforms genetic similarity, achieving 93.0\% vs. 73.2\% NDCG@1 and 98.4\% vs. 90.0\% overall NDCG.}

We train each ranker on CER or WER scores and find that choice of metric impacts ranking fidelity. For VAANI-D, the CER-based ranker performs better with a higher mean NDCG (98.4\%) than the WER-based ranker. A paired $t$-test across the overall NDCG scores for target languages shows that this difference is statistically significant ($t(17)=4.02$, $p<0.001$). One possible explanation for this result is that variation in spelling and word boundaries among the VAANI-D language varieties makes character-level evaluation more informative than word-level evaluation.

% In contrast, the WER-based ranker performs better on WAXAL (94.8\% versus 93.2\%), but the difference isn't significant.
% (paired t-test results $t=0.5956$, $p = 0.5588$). 
On WAXAL, genetic similarity remains a stronger baseline, particularly for top-ranked donors (Table 1), though DonorRank achieves a high overall NDCG. This shows that a learned ranking model isn't uniformly preferable to simpler donor-selection heuristics across multilingual settings. We see this reflected in how Amharic (amh) is the best performing donor for Tigrinya (tir), since they are the only two Semitic languages in WAXAL and share the Ge'ez script (see Table \ref{tab:waxal-donor-comparison}). 
Although both ranking objectives produce highly accurate rankings ($>90\%$ NDCG), this difference suggests that the most informative signal may depend on the characteristics of dataset. We use the ranker with a higher NDCG (CER-based for VAANI-D and WER-based for WAXAL) for subsequent experiments.

\subsection{DonorRank generalizes with limited transfer observations.}
\begin{table}[htbp]
    \centering \small
    \setlength{\tabcolsep}{3pt} % Tighten padding for single ACL column fit
    \begin{tabularx}{\columnwidth}{l *{4}{>{\centering\arraybackslash}X}}
        \toprule
        & \multicolumn{2}{c}{\textbf{VAANI-D}} & \multicolumn{2}{c}{\textbf{WAXAL}} \\
        \cmidrule(lr){2-3} \cmidrule(lr){4-5}
        \textbf{Experiment Setting} & \textbf{WER} & \textbf{CER} & \textbf{WER} & \textbf{CER} \\ 
        \midrule
        % (A) Unseen Donors, 50 Pairs & 0.940 & \textbf{0.986} & 0.887 & 0.879 \\
        % (B) Unseen Donors, Full & 0.948 & 0.986 & \textbf{0.926} & 0.901 \\
        % (C) Multi-Seed (5 seeds), 50 rows   & 0.930 & 0.984 & 0.902 & 0.870 \\
        % (D) Multi-Split (5 splits), 50 rows & 0.934 & 0.978 & 0.908 & 0.894 \\
        % (E) Multi-Split (5 splits), Full    & \textbf{0.955} & 0.983 & 0.912 & \textbf{0.950} \\
        (A) Unseen Donors, 50 Pairs & 94.0 & \textbf{98.6} & 88.7 & 87.9 \\
        (B) Unseen Donors, Full & 94.8 & 98.6 & \textbf{92.6} & 90.1 \\
        (C) Multi-Seed (5 seeds), 50 rows   & 93.0 & 98.4 & 90.2 & 87.0 \\
        (D) Multi-Split (5 splits), 50 rows & 93.4 & 97.8 & 90.8 & 89.4 \\
        (E) Multi-Split (5 splits), Full    & \textbf{95.5} & 98.3 & 91.2 & \textbf{95.0} \\
        \bottomrule
    \end{tabularx}
    \caption{DonorRank performance under limited transfer supervision and held-out-donor evaluation. Entries report mean WER- and CER-based NDCG on VAANI-D and WAXAL. Settings A–B compare training with 50 donor–target pairs vs. all available training pairs, C repeats the 50-pair setting with five random seeds, and D–E evaluate five language splits using limited and full training data, respectively. Higher values indicate better ranking quality.}
    \label{tab:exp_settings_ndcg}
\end{table}
We also check if DonorRank can produce informative donor rankings with limited observations. Table~\ref{tab:exp_settings_ndcg} reports ranking performance under reduced training data settings and evaluations involving held-out donor languages. 

We hold out seven languages, considering each language in turn as the target and the remaining six as candidate donors. In all experiments, the training pairs do not contain any of the held-out languages. We first train DonorRank using (A) 50 of observed donor--target pairs, and (B) all available training pairs for all the remaining languages. To examine the effect of sampling, we repeat the 50-pair setting using (C) five different random seeds. Finally, we evaluate (D) 50-pair and (E) full-data training across five different language splits, aggregating results over the resulting 35 evaluations. 

Table~\ref{tab:exp_settings_ndcg} shows that \textbf{DonorRank maintains high ranking quality with only 50 observed transfer pairs, approaching the performance obtained using all available training pairs}. The repeated-seed and repeated-split experiments yield broadly comparable aggregate results.

Together, these findings show that DonorRank learns useful donor-selection patterns from few observed transfers and applies them to new candidate languages. Rather than requiring complete pairwise transfer results, the trained model uses available linguistic and dataset-level features to predict promising donors.

\subsection{Donor selection matters for zero-shot ASR performance.}

\begin{table}[t]
\centering
\resizebox{\linewidth}{!}{%
\setlength{\tabcolsep}{2pt}
\begin{tabular}{@{}llcc lcc cc@{}}
\toprule
& \multicolumn{3}{c}{\textbf{DonorRank (A)}} & \multicolumn{3}{c}{\textbf{Genetic (B)}} & \multicolumn{2}{c}{\textbf{Hindi (C)}} \\
\cmidrule(lr){2-4} \cmidrule(lr){5-7} \cmidrule(lr){8-9}
\textbf{Target} & \textbf{Donor} & \textbf{CER} & \textbf{WER} & \textbf{Donor} & \textbf{CER} & \textbf{WER} & \textbf{CER} & \textbf{WER} \\
\midrule
anp & hne & 28.3 & 62.8 & mai & \textbf{22.1} & \textbf{53.1} & 37.7 & 81.1 \\
awa & hne & \textbf{22.4} & \textbf{55.0} & hne & \textbf{22.4} & \textbf{55.0} & 34.3 & 81.2 \\
bgc & hne & \textbf{28.2} & \textbf{64.2} & hin & 32.2 & 70.5 & 32.2 & 70.5 \\
bhb & anp & 55.2 & 114.2 & hne & \textbf{48.4} & \textbf{91.2} & 51.6 & 94.1 \\
bho & hne & \textbf{21.0} & \textbf{52.2} & sdr & 42.2 & 87.2 & 30.0 & 71.4 \\
gbm & kfy & \textbf{35.2} & \textbf{64.8} & kfy & \textbf{35.2} & \textbf{64.8} & 40.1 & 86.5 \\
hin & gbm & \textbf{22.1} & \textbf{50.1} & bgc & 99.9 & 100 & - & - \\
hlb & hne & \textbf{37.2} & \textbf{66.2} & kok & 44.0 & 83.4 & 45.9 & 89.4 \\
hne & kok & 37.2 & 80.4 & awa & 99.8 & 100  & \textbf{32.3} & \textbf{76.1} \\
kfy & hne & \textbf{17.1} & \textbf{38.9} & gbm & 22.3 & 51.9 & 18.5 & 45.7 \\
kok & mag & 47.5 & 91.2 & mar & \textbf{31.1} & \textbf{67.7} & 44.7 & 97.3 \\
mag & hne & 26.1 & \textbf{55.8} & anp & \textbf{25.7} & 58.0 & 34.4 & 72.3 \\
mai & mag & \textbf{20.5} & \textbf{46.8} & anp & 21.8 & 49.8 & 32.8 & 71.8 \\
mar & mag & 41.2 & 90.8 & kok & \textbf{28.4} & \textbf{68.0} & 44.0 & 98.3 \\
mwr & raj & \textbf{24.1} & \textbf{47.2} & bhb & 100  & 100  & 37.0 & 71.8 \\
nep & gbm & 35.1 & \textbf{101.4} & kfy & \textbf{32.0} & 101.8 & 43.9 & 110.6  \\
raj & mwr & \textbf{21.5} & \textbf{46.4} & bhb & 100  & 100  & 35.9 & 73.5 \\
sjp & hne & \textbf{34.7} & \textbf{77.8} & hlb & 38.7 & 89.4 & 42.3 & 93.2 \\
\bottomrule
\end{tabular}%
}
\caption{Zero-shot ASR performance on VAANI-D using donors selected by DonorRank, genetic similarity, or the fixed Hindi baseline. CER and WER (\%) are reported for each target language after fine-tuning w2vBERT on the selected donor. Bold indicates the lowest error rate for each target and metric.}
\label{tab:donor-comparison-hindi-baseline}
\end{table}

\begin{table}[t]
\centering
\resizebox{\linewidth}{!}{%
\setlength{\tabcolsep}{2pt}
\begin{tabular}{@{}llcc lcc lcc@{}}
\toprule
& \multicolumn{3}{c}{\textbf{DonorRank (A)}} & \multicolumn{3}{c}{\textbf{Genetic (B)}} & \multicolumn{3}{c}{\textbf{Oromo (C)}} \\
\cmidrule(lr){2-4} \cmidrule(lr){5-7} \cmidrule(lr){8-10}
\textbf{Target} & \textbf{Donor} & \textbf{CER} & \textbf{WER} & \textbf{Donor} & \textbf{CER} & \textbf{WER}  & \textbf{CER} & \textbf{WER} \\
\midrule
ach & sna & 45.5 & 98.4 & mas & \textbf{44.2} & \textbf{93.8} & 60.0 & 99.3 \\
aka & ful & 57.8 & \textbf{95.0} & ewe & \textbf{44.8} & 96.4 & 66.6 & 98.5 \\
amh & tir & \textbf{38.7} & \textbf{86.9} & tir & \textbf{38.7} & \textbf{86.9} & 96.7 & 102.9 \\
dag & ful & 52.2 & 92.1 & dga & \textbf{47.7} & \textbf{89.5} & 64.7 & 98.6 \\
dga & kpo & 48.8 & 95.0 & dag & \textbf{41.1} & \textbf{83.9} & 64.2 & 98.2 \\
ewe & ful & 61.2 & 95.1 & kpo & \textbf{47.2} & \textbf{94.9} & 73.1 & 99.2 \\
ful & ach & \textbf{54.0} & 98.6 & ewe & 56.6 & 109.8 & 55.2 & \textbf{97.4} \\
kpo & ful & 69.6 & 100.9 & ewe & \textbf{53.7} & 108.1 & 79.9 & \textbf{99.8} \\
lin & sna & \textbf{29.9} & 85.2 & lug & 33.9 & \textbf{80.7} & 57.7 & 99.9 \\
lug & sna & 27.7 & 93.9 & nyn & \textbf{21.7} & \textbf{82.4} & 47.2 & 100.2 \\
mas & sna & \textbf{33.9} & \textbf{94.6} & ach & 48.5 & 123.3 & 53.3 & 101.4 \\
mlg & sna & \textbf{41.3} & \textbf{97.8} & ewe & 51.8 & 121.3 & 57.2 & 100.0 \\
nyn & sna & 32.0 & 91.4 & lug & \textbf{28.6} & \textbf{88.6} & 52.9 & 100.7 \\
orm & sna & 38.5 & 99.2 & sid & \textbf{25.5} & \textbf{90.9} & - & - \\
sid & sna & 30.5 & 95.1 & orm & \textbf{25.7} & \textbf{94.3} & \textbf{25.7} & \textbf{94.3} \\
sna & lug & \textbf{32.3} & 114.7 & lug & \textbf{32.3} & 114.7 & 52.8 & \textbf{105.6} \\
sog & sna & \textbf{30.7} & \textbf{95.2} & ewe & 51.8 & 151.4 & 51.0 & 101.1 \\
tir & sna & 98.3 & 100.4 & amh & \textbf{62.3} & \textbf{100.0} & 96.6 & 103.2 \\
wal & sna & 35.5 & 98.1 & ewe & 49.1 & 153.0 & \textbf{31.5} & \textbf{94.5} \\
\bottomrule
\end{tabular}%
}
\caption{Zero-shot ASR performance on WAXAL using donors selected by DonorRank, genetic similarity, or the fixed Oromo baseline. CER and WER (\%) are reported for each target language after fine-tuning w2vBERT on the selected donor. Bold indicates the lowest error rate for each target and metric.}
\label{tab:waxal-donor-comparison}
\end{table}

We evaluate whether DonorRank’s top-ranked donor yields competitive zero-shot ASR performance relative to two practical selection baselines: the closest genetically related donor and a fixed high-resource donor. We use Hindi as the high-resource baseline for VAANI-D and Oromo for WAXAL. Tables~\ref{tab:donor-comparison-hindi-baseline} and \ref{tab:waxal-donor-comparison} report CER and WER after fine-tuning w2vBERT on the selected donor and evaluating on each target language.

On VAANI-D, DonorRank frequently identifies donors that outperform genetic similarity and the Hindi baseline, although the advantage varies across targets (Table~\ref{tab:donor-comparison-hindi-baseline}). We see that the DonorRank selected language beats the genetic baseline for 11 out of 18 target languages. \textbf{On VAANI-D, DonorRank reduces mean CER from 47.01\% with genetic similarity to 30.81\%.} For Bhojpuri, selecting Chhattisgarhi rather than the genetically closest donor reduces CER from 42.2\% to 21.0\% and WER from 87.2\% to 52.2\%. Similar improvements are observed for Haryanvi and Kumaoni. However, genetic similarity remains more effective for some targets, including Angika, Bhili, Konkani, and Marwari. These results show that learned donor selection can uncover beneficial transfer relationships beyond genealogical proximity, but doesn't consistently identify the best-performing donor.

On WAXAL, genetic similarity is a stronger competitor, outperforming DonorRank for several target languages (Table~\ref{tab:waxal-donor-comparison}). Nevertheless, DonorRank identifies useful alternatives in some cases. For Malagasy, its selected donor reduces CER from 51.8\% to 41.3\%, and for Maasai, CER decreases from 48.5\% to 33.9\%. These examples show that effective transfer can occur beyond the closest genetic neighbor, although the results also reveal limitations in ranking across a more heterogeneous collection of languages.

These results suggest that the role of donor ranking differs based on the donor set. While the median relative improvement in WER for VAANI-D is +7.48\%, on the WAXAL dataset DonorRank performs very close but slightly worse than the genetic baseline with an overall median relative improvement of -0.21\%. In VAANI-D, learned ranking helps distinguish effective donors among closely related candidate languages, often outperforming genetic similarity. In WAXAL, donor selection involves navigating a much broader multilingual landscape, where relatedness is a useful starting point but learned ranking can uncover more effective transfer partners. While donor selection can improve zero-shot transfer, high error rates for some targets highlight the limitations of donor-only zero-shot adaptation. Additional experiments combining top-ranked donors are reported in Appendix \ref{sec:multi_donor_experiments}.

\section{Decoding Donor Language Selection}
DonorRank also provides a framework for examining which linguistic and dataset-level features drive successful transfer across different settings.

\begin{figure}%[tbp]
    \centering
    \includegraphics[width=\columnwidth]{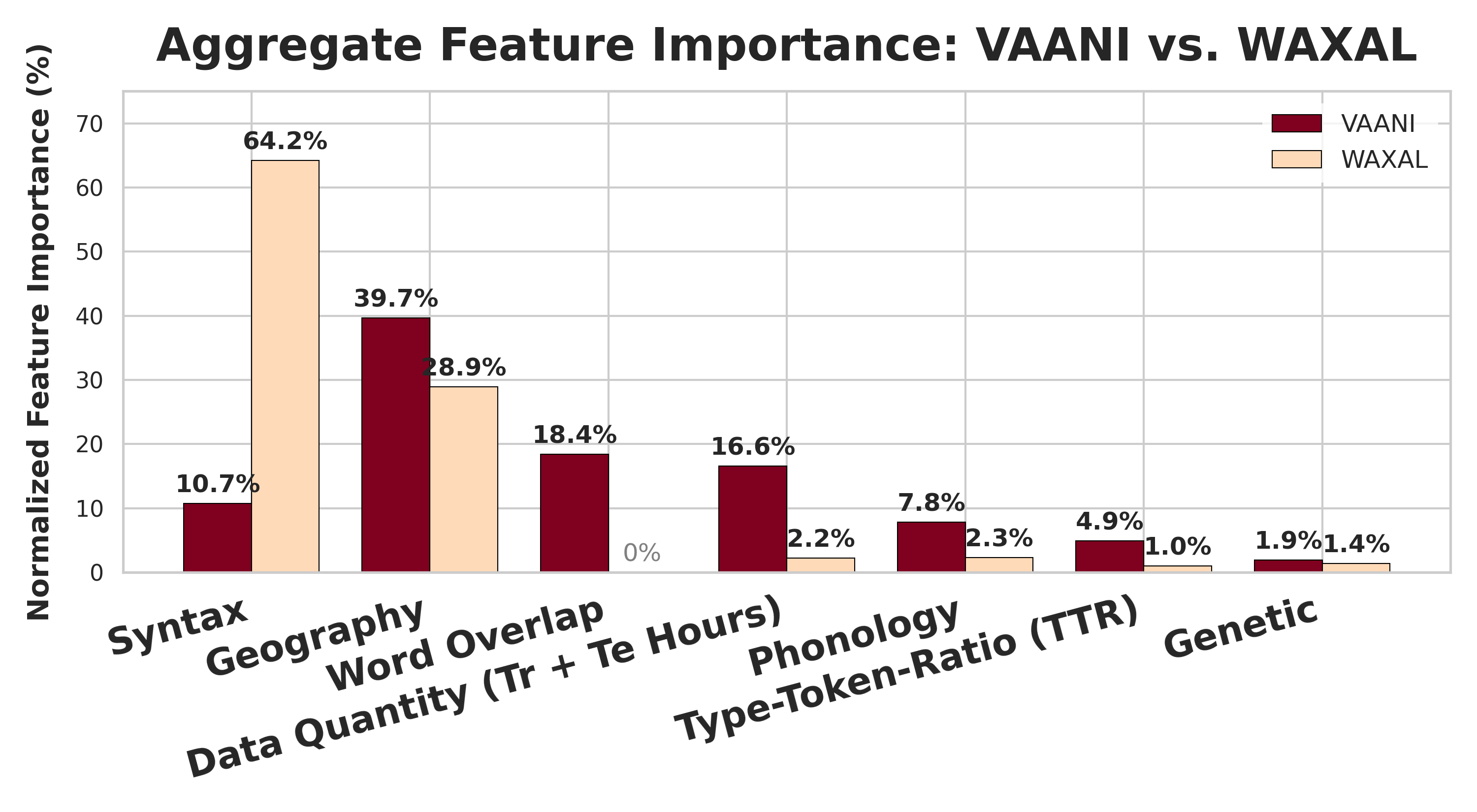}
    \caption{Normalized feature importance in DonorRank models trained on WAXAL and VAANI-D. Importance values indicate the relative contribution of each feature within the fitted ranking model.
    }
    \label{fig:feat_imp}
    \vspace{-4mm}
\end{figure}

% \subsection{What linguistic signals govern successful transfer?}

Figure~\ref{fig:feat_imp} compares feature importance across VAANI-D and WAXAL. For VAANI-D, geographic proximity, lexical overlap, and available training data are among the most influential features. These cues help discern donors among closely related languages that might also be mutually intelligible. In contrast, for WAXAL, syntactic and geographic features are more important. Because WAXAL has multiple language families and writing systems, structural linguistic properties are stronger indicators of transferability than lexical similarity alone. This suggests that \textbf{DonorRank learns different predictive signals depending on the linguistic composition of the candidate language set.} Notably, genetic similarity doesn't emerge as an important feature for either dataset, with normalized feature importance of 1.9\% and 1.4\% for VAANI and WAXAL, respectively.

\subsection{Phylogeny alone is not sufficient.}
Phylogenetic relatedness is a useful starting point for donor selection but isn't sufficient to explain transfer performance across either dataset.

In VAANI-D, several closely related language pairs, such as Awadhi (awa) -- Chhattisgarhi (hne) and Garhwali (gbm) -- Kumaoni (kfy), perform well under both DonorRank and genealogy-based selection. However, other cases show that the nearest linguistic relative isn't always the most effective donor. For example, DonorRank selects Chhattisgarhi (hne) rather than Hindi (hin) for Haryanvi (bgc), and Garhwali (gbm) rather than the genetically closest donor for Hindi (hin), yielding lower transcription error rates than the genetic baseline in both cases. These results show that \textbf{genealogical proximity alone does not guarantee effective transfer}.

WAXAL has a similar pattern where for some languages, DonorRank selects donors that differ from the nearest genealogical neighbor and yield improved transfer performance. For instance, Shona, a Bantu language, is the highest-ranked donor for Malagasy, an Austronesian language; these are completely unrelated languages phylogenetically, yet DonorRank's selection of Shona results in lower transcription error rates (e.g., 10.5-point CER reduction). These results show that genetic similarity is informative, but transferability is ultimately influenced by multiple linguistic and corpus-level factors.

\subsection{Successful transfer is characterized by complementary signals.}

\begin{table}[ht!]
\centering
\small
\setlength{\tabcolsep}{1.5pt}
\begin{tabular}{ccccccc cc}
\toprule
\multicolumn{7}{c}{\textbf{Features Included}} & \multicolumn{2}{c}{\textbf{Mean NDCG}} \\
\cmidrule(lr){1-7} \cmidrule(lr){8-9}
\textbf{Hrs} & \textbf{Phon} & \textbf{Gen} & \textbf{Syn} & \textbf{Geo} & \textbf{Ovlp} & \textbf{TTR} & \textbf{WAXAL} & \textbf{VAANI} \\
\midrule
\multicolumn{9}{l}{\textit{Single Features}} \\
\checkmark & & & & & \checkmark & \checkmark & 86.5 & -- \\
& \checkmark & & & & & & 87.0 & 95.2 \\
& & \checkmark & & & & & 93.5 & 94.5 \\
& & & \checkmark & & & & 93.7 & 91.2 \\
& & & & \checkmark & & & \textbf{95.3} & 95.0 \\
\midrule
\multicolumn{9}{l}{\textit{Multi-Feature Combinations}} \\
& \checkmark & \checkmark & \checkmark & & & & 93.8 & 95.7 \\
& \checkmark & & \checkmark & \checkmark & & & 94.6 & 95.8 \\
& & \checkmark & \checkmark & \checkmark & & & 95.1 & 96.0 \\
& \checkmark & \checkmark & & \checkmark & & & \textbf{95.3} & 96.3 \\
& \checkmark & \checkmark & \checkmark & \checkmark & & & 94.7 & 96.2 \\
% & \checkmark & \checkmark & \checkmark & \checkmark & \checkmark & \checkmark & 94.4 & -- \\
\midrule
\multicolumn{9}{l}{\textit{Full Dataset Models}} \\
\checkmark & \checkmark & \checkmark & \checkmark & \checkmark & \checkmark & \checkmark & 94.8 & \textbf{98.4} \\
% \checkmark & \checkmark & & \checkmark & \checkmark & \checkmark & \checkmark & -- & 98.5 \\
% \checkmark & \checkmark & \checkmark & \checkmark & & \checkmark & \checkmark & -- & 98.5 \\
\bottomrule
\end{tabular}
\caption{Feature-subset ablation of DonorRank on WAXAL and VAANI-D. Entries report mean NDCG when the ranking model is trained using the indicated linguistic and dataset-level features. Higher values indicate better ranking quality. Hrs=Training and Testing Hours, Phon=Phonological, Gen=Genetic, Syn=Syntactic, Geo=Geographical, Ovlp=Overlap Percentage, TTR=Type-to-Token Ratio.}
\label{tab:ablation_waxal_vaani_combined}
\vspace{-3mm}
\end{table}

We also perform ablation experiments on DonorRank using different subsets of linguistic and resource features (Table~\ref{tab:ablation_waxal_vaani_combined}). No single feature fully explains transfer, although several individual features produce strong rankings. In VAANI-D, phonological features provide the best performance (95.2\%), followed by geographic (95.0\%) and genetic features (94.5\%). In WAXAL, geographic features are the strongest individual predictor (95.3\%), while syntactic and genetic features also perform well (93.7 and 93.5\%). The strongest cue therefore differs across the two collections, while multiple forms of linguistic similarity are informative in both.

\textbf{Combining linguistic features improves ranking performance beyond most individual features. In VAANI-D, the best linguistic feature combination achieves 96.3\% NDCG, while the full model reaches 98.4\%.} In WAXAL, several linguistic combinations closely match the strongest individual geographic model. The full model remains competitive at 94.8\%, but doesn't outperform the best reduced configurations.

The performance of reduced-feature models shows that DonorRank can be adapted to settings where only a subset of linguistic information is available and the value of combining linguistic and dataset-level features depends on the language collection. On VAANI-D, the full feature set yields the highest ranking performance, whereas on WAXAL, smaller feature subsets perform comparably or slightly better. These results suggest that the most useful feature combinations depend on the candidate language collection.
\subsection{DonorRank finds potential transfer hubs.}
\begin{figure}[h]
    \centering
    \includegraphics[width=0.8\columnwidth]{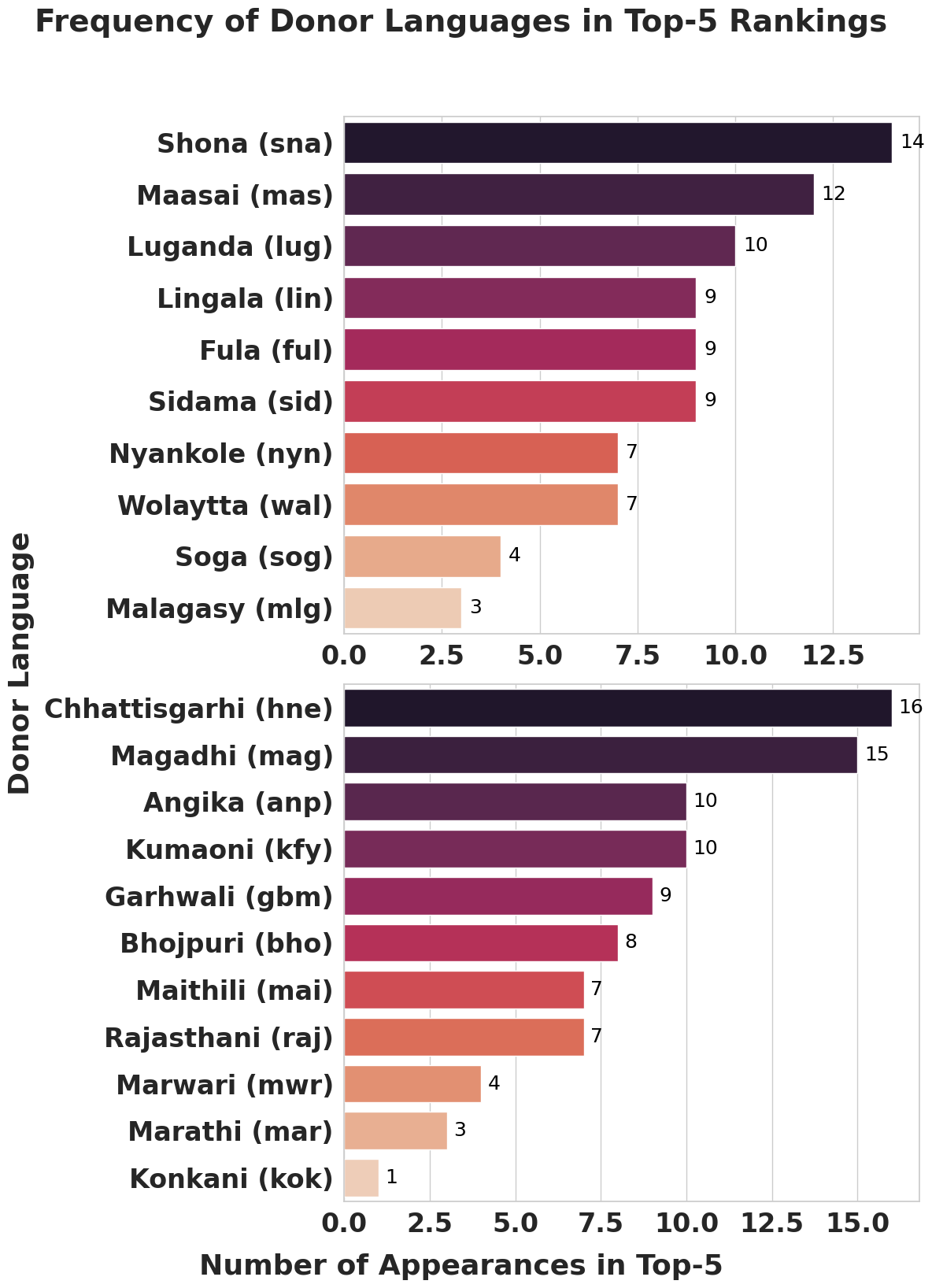}
    \caption{Most frequently appearing languages among DonorRank’s top-five predicted donors across WAXAL and VAANI-D.
    }
    \label{fig:top-5_distro}
    \vspace{-5mm}
\end{figure}

We also see that some languages appear frequently among DonorRank's highest-ranked donors across multiple targets (Figure 3).  Chhattisgarhi (hne) and Magadhi (mag) appear prominently in VAANI-D and Shona (sna), Maasai (mas), and Luganda (lug) in WAXAL. This suggests that certain languages may serve as promising transfer hubs within a multilingual collection, although their effectiveness depends on the target language. Some languages occupy central positions within the transfer landscape of a multilingual collection and identifying these hubs offers practical guidance for expanding multilingual ASR systems to new low-resource languages, when prior transfer experiments are unavailable.

\section{Conclusion}

We presented DonorRank, a learning-to-rank framework for donor language selection in low-resource cross-lingual ASR. \textbf{In the Indic Devanagari setting (VAANI-D), DonorRank substantially outperforms genetic similarity in donor ranking and zero-shot ASR performance, achieving high ranking quality even with limited transfer observations.} Our analyses further show that the usefulness of individual features and their combinations varies across multilingual settings, and that genetic similarity alone doesn't always identify the most effective transfer partner. DonorRank thus offers a practical approach to prioritizing candidate donors for new low-resource target languages, while providing a basis for analyzing transfer patterns across multilingual collections.

\section*{Limitations}
Our experiments focus on Indic and African languages and may not generalize to all language families or multilingual settings. We evaluate donor selection using a fixed w2vBERT fine-tuning setup; donor rankings may differ with other pretrained models, training strategies, or data conditions. Training DonorRank requires observed transfer outcomes, although our reduced-data experiments suggest that useful rankings can be learned from relatively few such observations. Finally, incomplete linguistic feature coverage and limited target-language text resources may affect ranking quality, particularly for poorly documented languages.

\section*{Ethical Considerations}
Our work aims to support automatic speech recognition (ASR) for low-resource languages, which are often underrepresented in speech technologies. By improving donor language selection, DonorRank may help reduce the computational cost of developing ASR systems for these languages. However, transfer performance varies across languages, and inaccurate transcriptions may disproportionately affect speakers of poorly resourced varieties. DonorRank also relies on linguistic and dataset-level features whose availability and quality may differ across languages, potentially reinforcing existing disparities in language coverage. Finally, donor selection should not substitute for community involvement in developing and evaluating language technologies. We encourage careful evaluation of ASR performance and responsible use of speech data, with attention to speaker privacy and the needs of the communities represented.

\bibliography{mybib}
\appendix
\section{Appendix}
\subsection{Saturation Effect in Multiple Donors}
\label{sec:multi_donor_experiments}

In addition to the single donor experiments covered in this paper, we also explore whether multiple donors selected by DonorRank can be utilized for effective ASR transfer. Although practitioners sometimes fine-tune on several related languages, the effect of increasing the number of donors is not well understood. We therefore conduct a systematic study of top-$k$ donor selection, analyzing how ASR performance evolves as additional ranked languages are included. For each target language, we use the trained ranking model to select the top-$k$ donor languages. We then fine-tune ASR models on the top-k donors (varying $k$ from 1 to 5) and evaluate zero-shot performance on the target. Among the VAANI-D languages, Awadhi (awa), Bhili (bhb), Garhwali (gbm), Halbi (hlb), Konkani (kok), and Marwari (mwr) serve as the primary low-resource target languages for evaluating zero-shot transfer using top-k languages.

To compare values of $k$ fairly in our multi-donor top-$k$ experiments, we balance total training data with per-donor allocation. For $k>1$, we allocate $\min(10/k,\, TrHr(d))$ hours per donor, where $TrHr(d)$ is the available training data for language $d$. For $k=1$, we cap training at 7 hours to avoid an imbalance with lower-resourced languages.

\begin{figure*}[h]
    \centering
    \includegraphics[width=\textwidth]{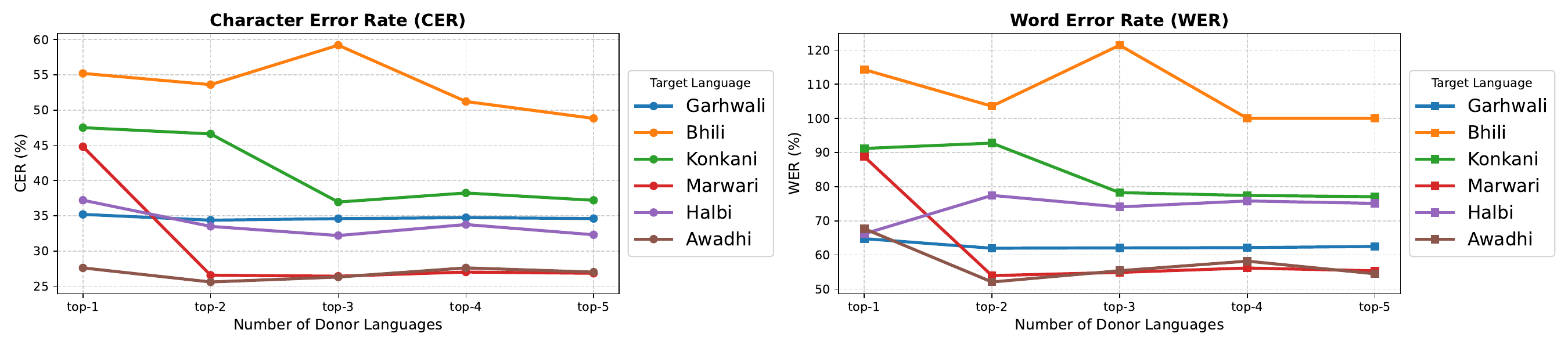}
    \caption{Performance of w2vBERT fine-tuned on top-k ranked donors for multiple target varieties. We find a negative correlation between CER and $k$ (mean Pearson's $\rho = -0.531$).}
    \label{fig:cer_wer_plot2}
\end{figure*}

In Figure~\ref{fig:cer_wer_plot2}, we see that across all six target languages, incorporating multiple donors generally improves performance over single-donor transfer. We find that the highest-ranked languages tend to capture most of the transferable information ($k=2...4$), while additional donors provide diminishing returns. This suggests a saturation effect: combining a small number of well-chosen donors is beneficial, but adding more languages yields limited gains and can introduce noise (as in the case of Bhili). These results show that DonorRank is also useful for constructing effective donor sets for multilingual adaptation. These findings have practical implications for low-resource ASR and dialectal settings, where variation can make indiscriminate aggregation of data harmful.

\subsection{DonorRank Cross Validation results on VAANI-D and WAXAL}

\begin{table*}
\centering
\small
\caption{Train/test hours (Tr/Te, capped at 7 hours), DonorRank top-5 donor languages, and NDCG scores across target languages, for VAANI-D (left) and WAXAL (right), using CER as the metric to train the ranker. 
% \aarohi{Do you have this table for the WER-based ranker for WAXAL? I think we should show those if all the other results in the table for WAXAL are based on the WER-based ranker.}
}
\label{tab:lgbm-rankings-cer-only}
\setlength{\tabcolsep}{3.5pt}
\renewcommand{\arraystretch}{0.92}
\begin{tabular}{@{}l c l c | l c l c@{}}
\toprule
\textbf{Target Language} & \textbf{Tr/Te} & \textbf{Top-5 Donors} & \textbf{NDCG} & \textbf{Target Language} & \textbf{Tr/Te} & \textbf{Top-5 Donors} & \textbf{NDCG} \\
\midrule
Angika (anp)        & 3.11 / 0.25 & hne, mag, mai, raj, bho & 98.9 & Akan (aka)         & 7.00 / 1.00 & sna, ful, lug, mas, kpo & 93.3 \\
Awadhi (awa)        & 0.24 / 0.02 & hne, mag, mwr, gbm, raj & 99.4 & Acholi (ach)       & 7.00 / 1.00 & ful, lug, sna, mas, sid & 99.5 \\
Bhili (bhb)         & 0.14 / 0.01 & anp, hne, kfy, mag, gbm & 94.4 & Amharic (amh)      & 7.00 / 1.00 & ful, sna, lug, sid, mas & 45.2 \\
Bhojpuri (bho)      & 7.00 / 1.00 & hne, mai, gbm, mag, anp & 99.6 & Dagbani (dag)      & 7.00 / 1.00 & ful, sna, mas, aka, dga & 97.1 \\
Chhattisgarhi (hne) & 7.00 / 1.00 & kok, mag, bho, gbm, anp & 97.5 & Dagaare (dga)      & 7.00 / 1.00 & kpo, dag, ewe, aka, ful & 98.1 \\
Garhwali (gbm)      & 7.00 / 0.62 & kfy, bho, hne, anp, mai & 99.0 & Ewe (ewe)          & 7.00 / 1.00 & aka, dga, ful, dag, sna & 96.5 \\
Halbi (hlb)         & 1.92 / 0.15 & hne, mag, mar, raj, kfy & 97.8 & Fulah (ful)        & 7.00 / 1.00 & dag, mas, sna, lug, dga & 98.1 \\
Haryanvi (bgc)      & 0.22 / 0.02 & hne, kfy, raj, bho, mai & 97.8 & Ikposo (kpo)       & 7.00 / 1.00 & dga, ful, ewe, dag, aka & 94.7 \\
Hindi (hin)         & 7.00 / 1.00 & gbm, kfy, mag, hne, anp & 98.6 & Lingala (lin)      & 7.00 / 1.00 & sna, lug, ful, sog, mas & 99.9 \\
Konkani (kok)       & 3.47 / 0.28 & mag, kfy, mar, raj, mwr & 95.9 & Ganda (lug)        & 7.00 / 1.00 & sna, ful, mas, sog, nyn & 97.8 \\
Kumaoni (kfy)       & 2.22 / 0.18 & hne, mwr, raj, mag, gbm & 99.0 & Masai (mas)        & 7.00 / 1.00 & ful, lug, sna, mlg, sid & 97.8 \\
Magadhi (mag)       & 5.64 / 0.45 & hne, mai, bho, anp, kfy & 99.6 & Malagasy (mlg)     & 7.00 / 1.00 & lug, sog, sna, mas, ful & 97.9 \\
Maithili (mai)      & 7.00 / 1.00 & mag, hne, bho, gbm, anp & 99.6 & Nyankole (nyn)     & 7.00 / 1.00 & sna, mas, lug, ful, sog & 99.3 \\
Marathi (mar)       & 7.00 / 1.00 & mag, kfy, anp, gbm, hne & 96.9 & Oromo (orm)        & 7.00 / 1.00 & mas, sna, ful, lug, sog & 96.1 \\
Marwari (mwr)       & 7.00 / 0.63 & raj, mag, hne, bho, mai & 99.9 & Sidamo (sid)       & 7.00 / 1.00 & lug, sna, ful, mas, orm & 97.4 \\
Nepali (nep)        & 7.00 / 1.00 & gbm, hne, mar, mag, kfy & 98.9 & Shona (sna)        & 7.00 / 1.00 & mas, lug, sog, ful, nyn & 99.7 \\
Rajasthani (raj)    & 7.00 / 0.85 & mwr, hne, kfy, mag, anp & 99.8 & Soga (sog)      & 7.00 / 1.00 & ful, sna, dag, lug, mlg & 94.5 \\
Surjapuri (sjp)     & 0.22 / 0.02 & hne, mai, mag, anp, bho & 99.0 & Tigrinya (tir)     & 7.00 / 1.00 & ful, sna, lug, ach, amh & 55.3 \\
                    &             &                         &       & Wolaytta (wal)     & 7.00 / 1.00 & lug, sna, ful, sid, mas & 97.3 \\
\midrule
\multicolumn{3}{@{}l}{\textbf{Mean NDCG}} & \textbf{98.4} & \multicolumn{3}{l}{\textbf{Mean NDCG}} & \textbf{93.2} \\
\bottomrule
\end{tabular}
\end{table*}

In Table \ref{tab:lgbm-rankings-cer-only}, we report the number of hours of training data and testing data available for each language. We then report the top-5 donors chosen by DonorRank and the overall NDCG score for each unseen target language held out for cross-validation.

\begin{table}
    \centering \footnotesize
    \setlength{\tabcolsep}{2pt}
    \resizebox{\columnwidth}{!}{%
    \begin{tabular}{l *{6}{c}}
        \toprule
        & \multicolumn{2}{c}{\textbf{\textit{DonorRank}}} & \multicolumn{2}{c}{\textbf{Random}} & \multicolumn{2}{c}{\textbf{Genetic}} \\
        \cmidrule(lr){2-3} \cmidrule(lr){4-5} \cmidrule(lr){6-7}
        \textbf{Dataset} & \textbf{WER} & \textbf{CER} & \textbf{WER} & \textbf{CER} & \textbf{WER} & \textbf{CER} \\ 
        \midrule
        \multicolumn{7}{c}{\textbf{\textit{Mean NDCG@1}}} \\
        \midrule
        WAXAL & 82.3$_{\scriptscriptstyle \!\pm\! 16.6}$ & 79.5$_{\scriptscriptstyle \!\pm\! 25.8}$ & 32.7$_{\scriptscriptstyle \!\pm\! 14.1}$ & 49.7$_{\scriptscriptstyle \!\pm\! 14.6}$ & \textbf{95.2}$_{\scriptscriptstyle \!\pm\! 12.4}$ & \textbf{98.1}$_{\scriptscriptstyle \!\pm\! 5.7}$ \\
        VAANI & \textbf{76.7}$_{\scriptscriptstyle \!\pm\! 22.1}$ & \textbf{93.0}$_{\scriptscriptstyle \!\pm\! 8.2}$ & 47.7$_{\scriptscriptstyle \!\pm\! 8.8}$ & 67.8$_{\scriptscriptstyle \!\pm\! 6.2}$ & 60.7$_{\scriptscriptstyle \!\pm\! 36.1}$ & 73.2$_{\scriptscriptstyle \!\pm\! 35.7}$ \\
        \midrule
        \multicolumn{7}{c}{\textbf{\textit{Mean NDCG@3}}} \\
        \midrule
        WAXAL & 85.7$_{\scriptscriptstyle \!\pm\! 14.1}$ & 82.9$_{\scriptscriptstyle \!\pm\! 24.9}$ & 44.8$_{\scriptscriptstyle \!\pm\! 13.2}$ & 58.4$_{\scriptscriptstyle \!\pm\! 14.7}$ & \textbf{92.2}$_{\scriptscriptstyle \!\pm\! 6.6}$ & \textbf{93.6}$_{\scriptscriptstyle \!\pm\! 6.9}$ \\
        VAANI & \textbf{80.2}$_{\scriptscriptstyle \!\pm\! 16.4}$ & \textbf{94.9}$_{\scriptscriptstyle \!\pm\! 4.9}$ & 51.6$_{\scriptscriptstyle \!\pm\! 5.7}$ & 69.6$_{\scriptscriptstyle \!\pm\! 3.6}$ & 61.0$_{\scriptscriptstyle \!\pm\! 24.4}$ & 71.5$_{\scriptscriptstyle \!\pm\! 28.3}$ \\
        \midrule
        \multicolumn{7}{c}{\textbf{\textit{Overall Mean NDCG}}} \\
        \midrule
        WAXAL & \textbf{94.8}$_{\scriptscriptstyle \!\pm\! 5.6}$ & 93.2$_{\scriptscriptstyle \!\pm\! 12.6}$ & 74.8$_{\scriptscriptstyle \!\pm\! 10.0}$ & 83.1$_{\scriptscriptstyle \!\pm\! 11.6}$ & 94.1$_{\scriptscriptstyle \!\pm\! 3.1}$ & \textbf{96.7}$_{\scriptscriptstyle \!\pm\! 2.3}$ \\
        VAANI & \textbf{92.6}$_{\scriptscriptstyle \!\pm\! 7.4}$ & \textbf{98.4}$_{\scriptscriptstyle \!\pm\! 1.5}$ & 80.2$_{\scriptscriptstyle \!\pm\! 2.9}$ & 88.6$_{\scriptscriptstyle \!\pm\! 1.3}$ & 84.4$_{\scriptscriptstyle \!\pm\! 8.5}$ & 90.0$_{\scriptscriptstyle \!\pm\! 7.9}$ \\
        \bottomrule
    \end{tabular}%
    }
    \caption{Mean NDCG evaluation scores (\%) with standard deviations in subscripts across DonorRank, Random Baseline, and Genetic Similarity on WAXAL and VAANI datasets. WER and CER indicate transfer-performance metrics. Best score per dataset and metric is in \textbf{bold}.}
    \label{tab:ndcg_scores_with_std_dev}
\end{table}

In Table \ref{tab:ndcg_scores_with_std_dev}, we report the aggregated overall NDCG, NDCG@1 and NDCG@3 scores for cross validation results across all target languages with the corresponding standard deviation. We also include the NDCG scores for a random ranking and a genetic ranking baseline. Across the VAANI dataset, we observe that the DonorRank framework outperforms the random and genetic baselines on predicting the overall ranking as well as the top-1 and top-3 donor, with limited variation for the CER-based ranker. On the other hand, the genetic baseline outperforms DonorRank on WAXAL on predicting the top-1 and top-3 donors. However, the WER based DonorRank is better at predicting overall rankings compared to the genetic baseline. There is also substantially more variation with higher standard deviation in the NDCG for the WAXAL dataset likely due to heterogeneity of the represented languages leading to variable transfer performance between language pairs.

\subsection{Feature Importance for DonorRank}

\begin{table}[ht]
\caption{Feature importance for CER-based (best) ranker for VAANI.} 
\label{tab:vaani-feature-importance}
\centering
\small
\begin{tabular}{@{}lr@{}}
\toprule
\textbf{Feature Name} & \textbf{Feature Importance} \\
\midrule
Donor Geographical & 279.3 \\
Overlap Percentage & 218.9 \\
Target Geographical & 194.3 \\
Donor Training hours & 143.4 \\
Target Syntactic & 78.9 \\
Target Type-to-Token Ratio & 58.4 \\
Target Testing hours & 54.7 \\
Target Phonological & 54.3 \\
Donor Syntactic & 48.3 \\
Donor Phonological & 39.1 \\
Donor Genetic & 13.8 \\
Target Genetic & 8.9 \\
\bottomrule
\end{tabular}
\end{table}
In Table \ref{tab:vaani-feature-importance}, we list the cumulative feature importance for both the CER-based ranker. Geographical features of the donor language emerge as the most important factor in determining successful donor transfer.

\begin{table}[ht]
\caption{Feature importance for WER-based (best) ranker for WAXAL.} 
\label{tab:waxal-feature-importance}
\centering
\small
\begin{tabular}{@{}lr@{}}
\toprule
\textbf{Feature Name} & \textbf{Feature Importance} \\
\midrule
Donor Syntactic & 396.0 \\
Target Syntactic & 350.1 \\
Target Geographical & 186.1 \\
Donor Geographical & 150.6 \\
Training Hours & 25.6 \\
Target Phonological & 15.2 \\
Donor Genetic & 12.2 \\
Target Type-to-Token Ratio & 12.1 \\
Donor Phonological & 11.2 \\
Target Genetic & 3.7 \\
\bottomrule
\end{tabular}
\end{table}
We see syntactic features of both the target and donor language emerge as the most important features for the WER-based ranker on the WAXAL dataset in Table \ref{tab:waxal-feature-importance}. The geographical features are the second most important factor determining cross-lingual transfer. Dataset features emerge as less important than these linguistic features likely due to the small size and standardized number of hours of the training and testing data we are using in the low-resource setting.

\begin{table}[ht]
\centering
\caption{Top 20 features for CER-based ranker on the VAANI dataset. HM $=$ headmark, N $=$ noun, Neg. $=$ negation.}
\label{tab:feature-importance-cer}
\resizebox{\linewidth}{!}{%
\begin{tabular}{@{}llr | llr@{}}
\toprule
\textbf{Type} & \textbf{Feature} & \textbf{Imp.} & \textbf{Type} & \textbf{Feature Description} & \textbf{Imp.} \\
\midrule
Data & Word Overlap \% & 247.2 & Geo. & GC\_-67\_-30 & 11.7 \\
Data & Donor Train Hrs & 104.2 & Geo. & GC\_-83\_42 & 11.7 \\
Data & Target Test Hrs & 29.4 & Synt. & Neg. After Subj. & 10.7 \\
Geo. & GC\_-83\_42 & 19.8 & Synt. & S\_SOV & 10.6 \\
Data & Type-To-Token & 19.4 & Synt. & Gender Mark & 10.5 \\
Synt. & Possessive HM & 15.7 & Synt. & Object HM & 10.1 \\
Geo. & GC\_-15\_-101 & 14.3 & Geo. & GC\_-51\_-38 & 9.3 \\
Geo. & GC\_-29\_-106 & 14.1 & Synt. & Neg. After Verb & 8.9 \\
Phon. & Alveolar Trill & 12.2 & Synt. & Possessor before N & 8.8 \\
Geo. & GC\_-67\_-30 & 11.7 & Gen. & Indo-Aryan Central & 8.1 \\
\bottomrule
\end{tabular}%
}
\end{table}

Table \ref{tab:feature-importance-cer} lists the most important individual features across the different feature groups for the CER-based DonorRank on the VAANI dataset. Lexical overlap between the target language and the donor language emerges as the most important factor, followed by the number of training and testing hours. We find that certain indices in the geographical vectors from lang2vec constantly rank amongst the most important features for determining successful cross-lingual transfer.

\end{document}